\pdfoutput=1
\documentclass[11pt]{article}

\usepackage{naacl2021}

\usepackage{times}
\usepackage{latexsym}

\usepackage[T1]{fontenc}
\usepackage[utf8]{inputenc}

\usepackage{microtype}

\usepackage{amsfonts}
\usepackage{amsmath}
\usepackage{booktabs}
\usepackage{cite}
\usepackage{multirow}
\usepackage{graphicx}

\usepackage{microtype}

\newcommand\wordnet{\textsc{WordNet}}
\newcommand\model{\textsc{CTP}}

\newif\ifcomments
\commentstrue 
\ifcomments
    \providecommand{\cc}[1]{{\protect\color{blue}{[cc: #1]}}}
    \providecommand{\kl}[1]{{\protect\color{green}{[kl: #1]}}}

\else
    \providecommand{\cc}[1]{}
    \providecommand{\kl}[1]{}
\fi

\newcommand\blfootnote[1]{%
  \begingroup
  \renewcommand\thefootnote{}\footnote{#1}%
  \addtocounter{footnote}{-1}%
  \endgroup
}

\title{Constructing Taxonomies from Pretrained Language Models}

\author{Catherine Chen$^*$  \hfill   Kevin Lin$^*$  \hfill   Dan Klein \\
  University of California, Berkeley \\
  {\tt \{cathychen,k-lin,klein\}@berkeley.edu}}
  
\begin{document}
\maketitle
\begin{abstract}
    We present a method for constructing taxonomic trees (e.g., \wordnet{}) using pretrained language models. Our approach is composed of two modules, one that \emph{predicts} parenthood relations and another that \emph{reconciles} those predictions into trees. The parenthood prediction module produces likelihood scores for each potential parent-child pair, creating a graph of parent-child relation scores. The tree reconciliation module treats the task as a graph optimization problem and outputs the maximum spanning tree of this graph.
    We train our model on subtrees sampled from \wordnet{}, and test on non-overlapping \wordnet{} subtrees. We show that incorporating web-retrieved glosses can further improve performance. On the task of constructing subtrees of English \wordnet{}, the model achieves 66.7 ancestor $F_1$, a 20.0\% relative increase over the previous best published result on this task. In addition, we convert the original English dataset into nine other languages using \textsc{Open Multilingual WordNet} and extend our results across these languages.

\end{abstract}
\blfootnote{* indicates equal contribution}
\section{Introduction}
 
A variety of NLP tasks use taxonomic information, including question answering \citep{miller1998wordnet} and information retrieval \citep{yang2012irwordnet}. Taxonomies are also used as a resource for building knowledge and systematicity into neural models \citep{peters2019knowledge,geiger2020modular,Talmor2020TeachingPM}. NLP systems often retrieve taxonomic information from lexical databases such as \wordnet~\citep{miller1998wordnet}, which consists of taxonomies that contain semantic relations across many domains. While manually curated taxonomies provide useful information, they are incomplete and expensive to maintain \citep{Hovy2009TowardCI}.

Traditionally, methods for automatic taxonomy construction have relied on statistics of web-scale corpora. These models generally apply lexico-syntactic patterns \citep{hearst1992automatic} to large corpora, and use corpus statistics to construct taxonomic trees \citep[e.g.,][]{snow2005learning,kozareva2010semi,bansal2014structured,mao-etal-2018-end,shang2020taxonomy}. 

\begin{figure}[ht]
\centering
\includegraphics[width=\linewidth
]{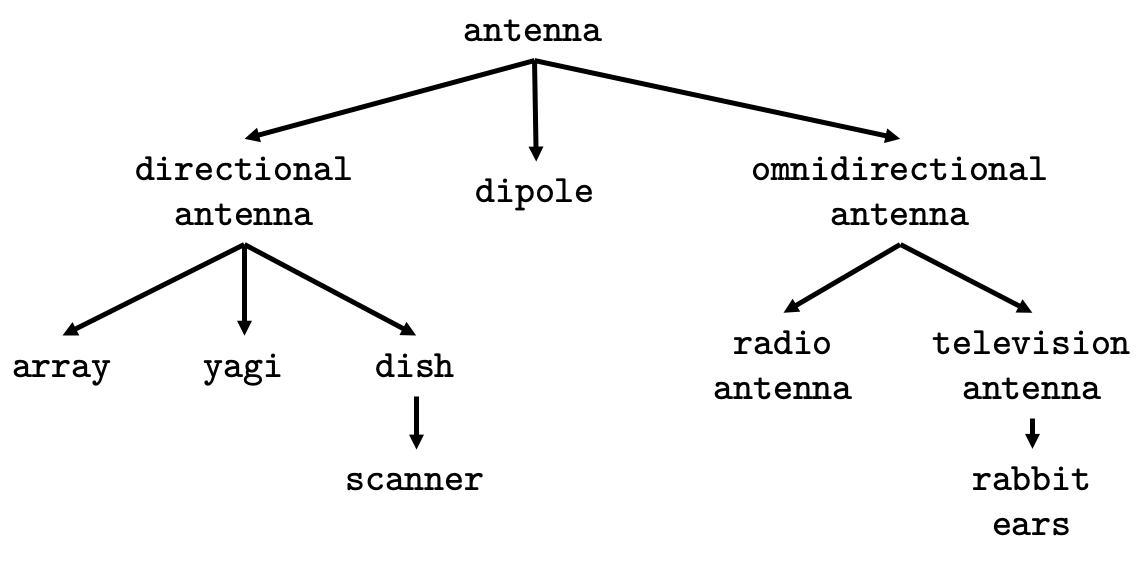}
\vspace{-0.43cm}
\caption{\textit{An example subtree from the \wordnet~hierarchy.}}
\label{fig:wordnet_example}
\end{figure}

% However, recent work has shown that some relational world knowledge can be extracted from pretrained language models \citep{bouraoui2019inducing, Bosselut2019COMETCT}.

In this work, we propose an approach that \textbf{c}onstructs \textbf{t}axonomic trees using \textbf{p}retrained language models (\model). Our results show that direct access to corpus statistics at test time is not necessary. Indeed, the re-representation latent in large-scale models of such corpora can be beneficial in constructing taxonomies. We focus on the task proposed by \citet{bansal2014structured}, where the task is to organize a set of input terms into a taxonomic tree. We convert this dataset into nine other languages using synset alignments collected in \textsc{Open Multilingual Wordnet} and evaluate our approach in these languages.

\model~first finetunes pretrained language models to \emph{predict} the likelihood of pairwise parent-child relations, producing a graph of parenthood scores. Then it \emph{reconciles} these predictions with a maximum spanning tree algorithm, creating a tree-structured taxonomy. We further test \model~in a setting where models have access to web-retrieved glosses. We reorder the glosses and finetune the model on the reordered glosses in the parenthood prediction module.

We compare model performance on subtrees across semantic categories and subtree depth, provide examples of taxonomic ambiguities, describe conditions for which retrieved glosses produce greater increases in tree construction $F_1$ score, and evaluate generalization to large taxonomic trees \citep{bordea2016}. These analyses suggest specific avenues of future improvements to automatic taxonomy construction.

Even without glosses, \model~achieves a 7.9 point absolute improvement in $F_1$ score on the task of constructing \wordnet~ subtrees, compared to previous work. When given access to the glosses, \model~obtains an additional 3.2 point absolute improvement in $F_1$ score. Overall, the best model achieves a 11.1 point absolute increase (a 20.0\% relative increase) in $F_1$ score over the previous best published results on this task.

Our paper is structured as follows. In Section \ref{sec:methods} we describe \model, our approach for taxonomy construction. In Section \ref{sec:experiments} we describe the experimental setup, and in Section \ref{sec:results} we present the results for various languages, pretrained models, and glosses. In Section \ref{sec:analysis} we analyze our approach and suggest specific avenues for future improvement. We discuss related work and conclude in Sections \ref{sec:related_work} and \ref{sec:discussion}.

\section{Constructing Taxonomies from Pretrained Models}\label{sec:methods}
\subsection{Taxonomy Construction}
We define taxonomy construction as the task of creating a tree-structured hierarchy $T=(V,E)$, where $V$ is a set of terms and $E$ is a set of directed edges representing hypernym relations. In this task, the model receives a set of terms $V$, where each term can be a single word or a short phrase, and it must construct the tree $T$ given these terms. \model{} performs taxonomy construction in two steps: parenthood prediction (Section \ref{sec:parenthood_prediction}) followed by graph reconciliation (Section \ref{sec:graph_pruning}).

\begin{figure*}[ht]
\centering
\includegraphics[width=0.9\textwidth
]{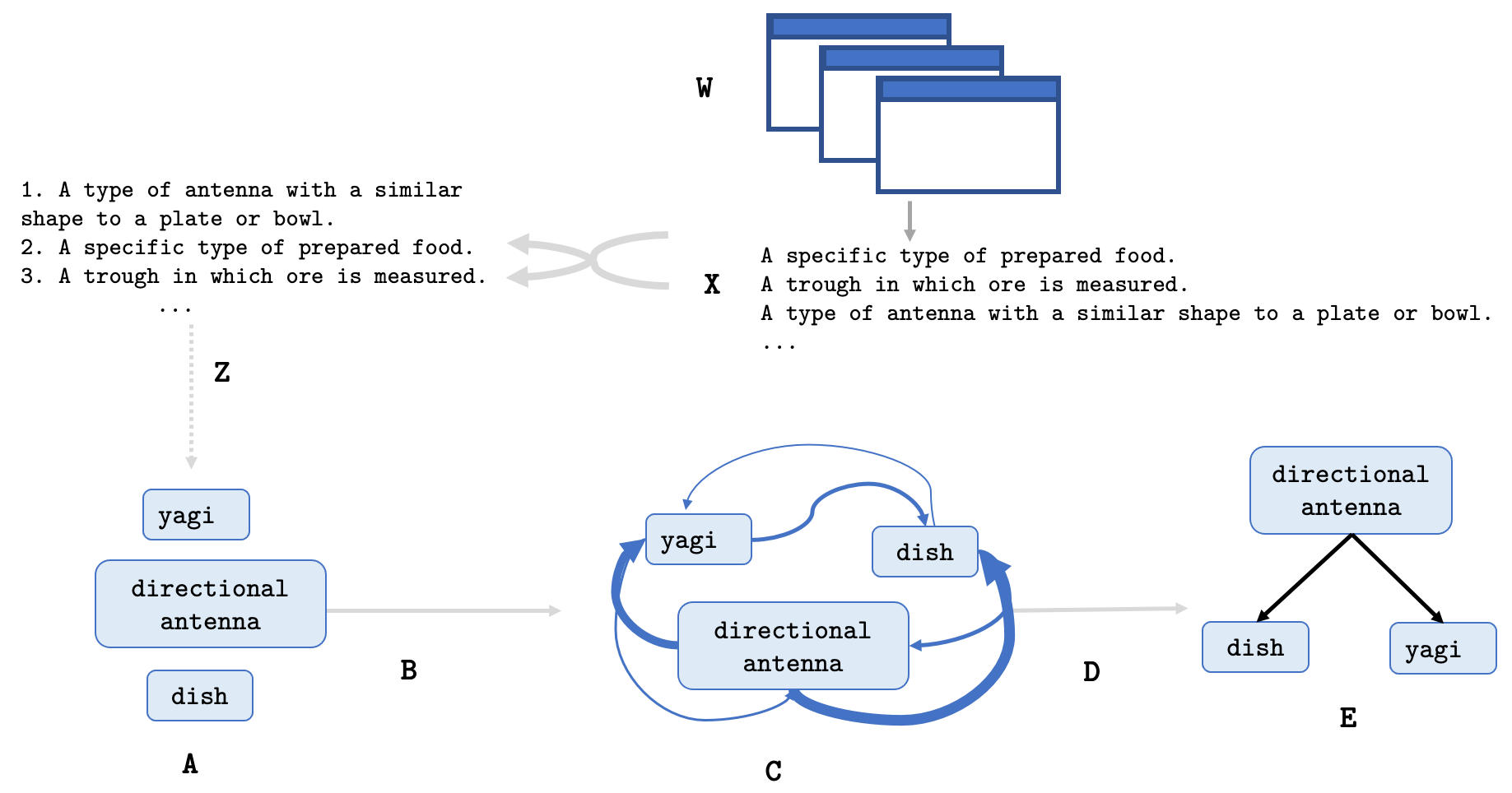}
\vspace{-0.43cm}
\caption{\textit{A schematic depiction of \model{}.} We start with a set of terms (A). We fine-tune a pretrained language model to predict pairwise parenthood relations between pairs of terms (B), creating a graph of parenthood predictions (C) (Section \ref{sec:parenthood_prediction}). We then reconcile the edges of this graph into a taxonomic tree (E) (Section \ref{sec:graph_pruning}). Optionally, we provide the model ranked web-retrieved glosses (Section \ref{sec:web_definitions}). We re-order the glosses based on relevance to the current subtree (Z).}
\label{fig:approach}
\end{figure*}

We provide a schematic description of \model{} in Figure \ref{fig:approach} and provide details in the remainder of this section.

\subsection{Parenthood Prediction}\label{sec:parenthood_prediction}
We use pretrained models (e.g., \textsc{BERT}) to predict the edge indicators $\mathbb{I}[parent(v_i,v_j)]$, which denote whether $v_i$ is a parent of $v_j$, for all pairs $(v_i,v_j)$ in the set of terms $V=\{v_1,...,v_n\}$ for each subtree $T$. 

To generate training data from a tree $T$ with $n$ nodes, we create a positive training example for each of the $n-1$ parenthood edges and a negative training example for each of the $\frac{n(n-1)}{2}-(n-1)$ pairs of nodes that are not connected by a parenthood edge.

We construct an input for each example using the template \texttt{$v_i$ is a $v_j$}, e.g., ``A dog is a mammal." Different templates (e.g., \texttt{[TERM\_A] is an example of [TERM\_B]} or \texttt{[TERM\_A] is a type of [TERM\_B]}) did not substantially affect model performance in initial experiments, so we use a single template. The inputs and outputs are modeled in the standard format \citep{Devlin2019BERTPO}.

We fine-tune pretrained models to predict $\mathbb{I}[parent(v_i,v_j)]$, which indicates whether $v_i$ is the parent of $v_j$, for each pair of terms using a sentence-level classification task on the input sequence.

\subsection{Tree Reconciliation}\label{sec:graph_pruning}
We then reconcile the parenthood graph into a valid tree-structured taxonomy. We apply the Chu-Liu-Edmonds algorithm to the graph of pairwise parenthood predictions. This algorithm finds the maximum weight spanning arborescence of a directed graph. It is the analog of MST for directed graphs, and finds the highest scoring arborescence in $O(n^2)$ time \citep{chu1965shortest}.

\subsection{Web-Retrieved Glosses}\label{sec:web_definitions}
We perform experiments in two settings: with and without web-retrieved glosses.
In the setting without glosses, the model performs taxonomy construction using only the set of terms $V$. In the setting with glosses, the model is provided with glosses retrieved from the web. For settings in which the model receives glosses, we retrieve a list of glosses $d_v^1,...,d_v^n$ for each term  $v\in V$.\footnote{We scrape glosses from wiktionary.com, merriam-webster.com, and wikipedia.org. For wikitionary.com and merriam-webster.com we retrieve a list of glosses from each site. For wikipedia.org we treat the first paragraph of the page associated with the term as a single gloss. The glosses were scraped in August 2020.}

Many of the terms in our dataset are polysemous, and the glosses contain multiple senses of the word. For example, the term \underline{dish} appears in the subtree we show in Figure \ref{fig:wordnet_example}. The glosses for \underline{dish} include (1) \textit{(telecommunications) A type of antenna with a similar shape to a plate or bowl}, (2) \textit{(metonymically) A specific type of prepared food}, and (3) \textit{(mining) A trough in which ore is measured}.

We reorder the glosses based on their relevance to the current subtree. We define relevance of a given context $d_v^i$ to subtree $T$ as the cosine similarity between the average of the GloVe embeddings \citep{pennington2014glove} of the words in $d_v^i$ (with stopwords removed), to the average of the GloVe embeddings of all terms $v_1,...,v_n$ in the subtree. This produces a reordered list of glosses $d_v^{(1)},...,d_v^{(n)}$.

We then use the input sequence containing the reordered glosses ``\texttt{[CLS]} $ v_i \: d_{v_i}^{(1)},...,d_{v_i}^{(n)}$. \texttt{[SEP]} $ v_j \: d_{v_j}^{(1)},...,d_{v_j}^{(n)}$” to fine-tune the pretrained models on pairs of terms $(v_i, v_j)$.
\section{Experiments}\label{sec:experiments}
In this section we describe the details of our datasets (Section \ref{sec:datasets}), and describe our evaluation metrics (Section \ref{sec:evaluation}). We ran our experiments on a cluster with 10 Quadro RTX 6000 GPUs. Each training runs finishes within one day on a single GPU.

\subsection{Datasets}\label{sec:datasets}
We evaluate \model{} using the dataset of medium-sized \wordnet~subtrees created by \citet{bansal2014structured}. This dataset consists of bottomed-out full subtrees of height 3 (this corresponds to trees containing 4 nodes in the longest path from the root to any leaf) that contain between 10 and 50 terms. This dataset comprises 761 English trees, with 533/114/114 train/dev/test trees respectively.

\subsubsection{Multilingual \wordnet}\label{sec:multilingual_dataset}
\wordnet~was originally constructed in English, and has since been extended to many other languages such as Finnish \citep{Magnini1994APF}, Italian \citep{Lindn2014IsIP}, and Chinese \citep{Wang2013BuildingTC}. Researchers have provided alignments from synsets in English \wordnet~to terms in other languages, using a mix of automatic and manual methods \citep[e.g.,][]{Magnini1994APF,Lindn2014IsIP}. These multilingual wordnets are collected in the \textsc{Open Multilingual WordNet} project \citep{bondpaik_2012}. The coverage of synset alignments varies widely. For instance, the alignment of \textsc{Albanet} (Albanian) to English \wordnet~ covers 3.6\% of the synsets in the \citet{bansal2014structured} dataset, while the \textsc{FinnWordNet} (Finnish) alignment covers 99.6\% of the synsets in the dataset.

We convert the original English dataset to nine other languages using the synset alignments. (We create datasets for Catalan \citep{Agirre2011MultilingualCR}, Chinese \citep{Wang2013BuildingTC}, Finnish \citep{Lindn2014IsIP}, French \citep{Sagot2008BuildingAF}, Italian \citep{Magnini1994APF}, Dutch \citep{Postma:Miltenburg:Segers:Schoen:Vossen:2016}, Polish \citep{Piasecki2009AWF}, Portuguese \citep{Paiva2012RevisitingAB}, and Spanish \citep{Agirre2011MultilingualCR}).

Since these wordnets do not include alignments to all of the synsets in the English dataset, we convert the English dataset to each target language using alignments specified in \wordnet~as follows. We first exclude all subtrees whose roots are not included in the alignment between the \wordnet~of the target language and English \wordnet. For each remaining subtree, we remove any node that is not included in the alignment. Then we remove all remaining nodes that are no longer connected to the root of the corresponding subtrees. We describe the resulting dataset statistics in Table \ref{tab:multilingual_dataset} in the Appendix.

\subsection{Evaluation Metrics}\label{sec:evaluation}
As with previous work \citep{bansal2014structured,mao-etal-2018-end}, we report the ancestor $F_1$ score $\frac{2 P R}{P + R}$, where

\begin{align*}
P&=\frac{|\mbox{\textsc{is\_a}}_{\mbox{\textsc{predicted}}} \cap \mbox{\textsc{is\_a}}_{\mbox{\textsc{gold}}}|}{| \mbox{\textsc{is\_a}}_{\mbox{\textsc{predicted}}}|}\\
R&=\frac{|\mbox{\textsc{is\_a}}_{\mbox{\textsc{predicted}}} \cap \mbox{\textsc{is\_a}}_{\mbox{\textsc{gold}}}|}{|\mbox{\textsc{is\_a}}_{\mbox{\textsc{gold}}}|}
\end{align*}

$\mbox{\textsc{is\_a}}_{\mbox{\textsc{predicted}}}$ and $\mbox{\textsc{is\_a}}_{\mbox{\textsc{gold}}}$ denote the set of predicted and gold ancestor relations, respectively.
We report the mean precision ($P$), recall ($R$), and $F_1$ score, averaged across the subtrees in the test set.

\subsection{Models}
In our experiments, we use pretrained models from the Huggingface library \citep{wolf2019huggingface}. For the English dataset we experiment with \textsc{BERT}, \textsc{BERT}-Large, and \textsc{RoBERTa}-Large in the parenthood prediction module. We experiment with multilingual \textsc{BERT} and language-specific pretrained models (detailed in Section \ref{sec:mling_models_list} in the Appendix). We finetuned each model using three learning rates \{1e-5, 1e-6, 1e-7\}. For each model, we ran three trials using the learning rate that achieved the highest dev $F_1$ score. In Section \ref{sec:results}, we report the average scores over three trials. We include full results in Tables~\ref{tab:en_lr} and \ref{tab:multilingual_lr} in the Appendix. The code and datasets are available at \url{https://github.com/cchen23/ctp}.

\section{Results}\label{sec:results}
\subsection{Main Results}
Our approach, \model{}, outperforms existing state-of-the-art models on the \wordnet~ subtree construction task. In Table \ref{tab:english} we provide a comparison of our results to previous work. Even without retrieved glosses, \model{} with \textsc{RoBERTa-Large} in the parenthood prediction module achieves higher $F_1$ than previously published work. \model{} achieves additional improvements when provided with the web-retrieved glosses described in Section \ref{sec:web_definitions}.

We compare different pretrained models for the parenthood prediction module, and provide these comparisons in Section \ref{sec:model_comparison}.

\begin{table}[!hbt]
\centering
%\columnwidth
\begin{tabular}{lccc}
\toprule 
                & \textbf{P}  & \textbf{R}  & \textbf{F1} \\  \midrule

\citet{bansal2014structured}      &    48.0        &     55.2        &  51.4 \\ 
\citet{mao-etal-2018-end}   &     52.9       &      58.6       &  55.6 \\ \midrule

\model~(no glosses) & 67.3 & 62.0 & 63.5 \\\midrule

\model~(web glosses) & \textbf{69.3} & \textbf{66.2} & \textbf{66.7} \\ \bottomrule

\end{tabular}
\caption{\textit{English Results, Comparison to Previous Work}. Our approach outperforms previous approaches on reconstructing \wordnet~subtrees, even when the model is not given web-retrieved glosses.}\label{tab:english}
\end{table}

\subsection{Web-Retrieved Glosses}
In Table \ref{tab:defs} we show the improvement in taxonomy construction with two types of glosses -- glosses retrieved from the web (as described in Section \ref{sec:web_definitions}), and those obtained directly from \wordnet. We consider using the glosses from \wordnet~ as an oracle setting since these glosses are directly generated from the gold taxonomies. Thus, we focus on the web-retrieved glosses as the main setting. Models produce additional improvements when given \wordnet~ glosses. These improvements suggest that reducing the noise from web-retrieved glosses could improve automated taxonomy construction.
\begin{table}[!th]
\centering
%\columnwidth
\begin{tabular}{lccc}
\toprule 
                & \textbf{P}  & \textbf{R}  & \textbf{F1} \\  \midrule

\model~     & 67.3 & 62.0 & 63.5 \\\midrule

+ web glosses  & 69.3 & 66.2 & 66.7 \\\midrule

+ oracle glosses  & 84.0 & 83.8 & 83.2 \\\bottomrule

\end{tabular}
\caption{\textit{English Results, Gloss Comparison on Test Set.} 
Adding web glosses improves performance over only using input terms. Models achieve additional improvements in subtree reconstruction when given oracle glosses from \wordnet{}, showing possibilities for improvement in retrieving web glosses.}

\label{tab:defs}
\end{table}

\subsection{Comparison of Pretrained Models}\label{sec:model_comparison}

For both settings (with and without web-retrieved glosses), \model{} attains the highest $F_1$ score when \textsc{RoBERTa}-Large is used in the parenthood prediction step. As we show in Table \ref{tab:comparison}, the average $F_1$ score improves with both increased model size and with switching from \textsc{BERT} to \textsc{RoBERTa}.

\begin{table}[!th]
\centering
%\columnwidth
\begin{tabular}{lccc}
\toprule 
                & \textbf{P}  & \textbf{R}  & \textbf{F1} \\  \midrule

\model~(\textsc{BERT}-Base)      & 57.9 & 51.8 & 53.4 \\\midrule

\model~(\textsc{BERT}-Large)      & 65.5 & 59.8 & 61.4 \\\midrule

\model~(\textsc{RoBERTa}-Large)   & \textbf{67.3} & \textbf{62.0} & \textbf{63.5} \\\midrule

\end{tabular}
\caption{\textit{English Results, Comparison of Pretrained Models on Test Set.}
Larger models perform better and \textsc{RoBERTa} outperforms BERT.}\label{tab:comparison}
\end{table}

\subsection{Aligned Wordnets}
We extend our results to the nine non-English alignments to the \citet{bansal2014structured} dataset that we created. In Table \ref{tab:multilingual} we compare our best model in each language to a random baseline. We detail the random baseline in Section \ref{sec:random_baseline} in the Appendix and provide results from all tested models in Section \ref{tab:multilingual_full} in the Appendix.

\begin{table}[!th]
\centering
\begin{tabular}{p{0.25cm} lccc}
\toprule
     & Model & \textbf{P} & \textbf{R} & \textbf{F1} \\\midrule
    \multirow{2}{*}{ca} & Random Baseline & 20.0 & 31.3 & 23.6 \\
     & \model~(\textsc{mBERT}) & \textbf{38.7} & \textbf{39.7} & \textbf{38.0} \\\midrule
    \multirow{2}{*}{zh} & Random Baseline & 25.8 & 35.9 & 29.0 \\
     & \model~(\textsc{Chinese BERT}) & \textbf{62.2} & \textbf{57.3} & \textbf{58.7} \\\midrule
     \multirow{2}{*}{en} & Random Baseline & 8.9 & 22.2 & 12.4 \\
     & \model~(\textsc{RoBERTa}-Large) & \textbf{67.3} & \textbf{62.0} & \textbf{63.5} \\\midrule
    \multirow{2}{*}{fi} & Random Baseline & 10.1 & 22.5 & 13.5 \\
     & \model~(\textsc{FinBERT}) & \textbf{47.9} & \textbf{42.6} & \textbf{43.8} \\\midrule
    \multirow{2}{*}{fr} & Random Baseline & 22.1 & 34.4 & 25.9 \\
     & \model~(\textsc{French BERT}) & \textbf{51.3} & \textbf{49.1} & \textbf{49.1} \\\midrule
    \multirow{2}{*}{it} & Random Baseline & 28.9 & 39.4 & 32.3 \\
     & \model~(\textsc{Italian BERT}) & \textbf{48.3} & \textbf{45.5} & \textbf{46.1} \\\midrule
    \multirow{2}{*}{nl} & Random Baseline & 26.8 & 38.4 & 30.6 \\
     & \model~(\textsc{BERTje}) & \textbf{44.6} & \textbf{44.8} & \textbf{43.7} \\\midrule
    \multirow{2}{*}{pl} & Random Baseline & 23.4 & 33.6 & 26.8 \\
     & \model~(\textsc{Polbert}) & \textbf{51.9} & \textbf{49.7} & \textbf{49.5} \\\midrule
    \multirow{2}{*}{pt} & Random Baseline & 26.1 & 37.6 & 29.8 \\
     & \model~(\textsc{BERTimbau}) & \textbf{59.3} & \textbf{57.1} & \textbf{56.9} \\\midrule
    \multirow{2}{*}{es} & Random Baseline & 27.0 & 37.2 & 30.5 \\
     & \model~(\textsc{BETO}) & \textbf{53.1} & \textbf{51.7} & \textbf{51.7} \\\bottomrule
\end{tabular}
\caption{\textit{Multilingual \wordnet~Test Results.} We extend our model to datasets in nine other languages, and evaluate our approach on these datasets. We use ISO 639-1 acronyms to indicate languages.}\label{tab:multilingual}
\end{table}

\model's $F_1$ score non-English languages is substantially worse than its $F_1$ score on English trees. Lower $F_1$ scores in non-English languages are likely due to multiple factors. First, English pretrained language models generally perform better than models in other languages because of the additional resources devoted to the development of English models. \citep[See e.g.,][]{Bender2011OnAA,mielke2016,joshi2020state}. Second, \textsc{Open Multilingual Wordnet} aligns wordnets to English \wordnet, but the subtrees contained in English \wordnet~might not be the natural taxonomy in other languages. However, we note that scores across languages are not directly comparable as dataset size and coverage vary across languages (as we show in Table \ref{tab:multilingual_dataset}).

These results highlight the importance of evaluating on non-English languages, and the difference in available lexical resources between languages. Furthermore, they provide strong baselines for future work in constructing wordnets in different languages.

\section{Analysis}\label{sec:analysis}
In this section we analyze the models both quantitatively and qualitatively. Unless stated otherwise, we analyze our model on the dev set and use \textsc{RoBERTa}-Large in the parenthood prediction step.

\subsection{Models Predict Flatter Trees}

\begin{figure}[ht]
\centering
\includegraphics[width=0.93\linewidth
]{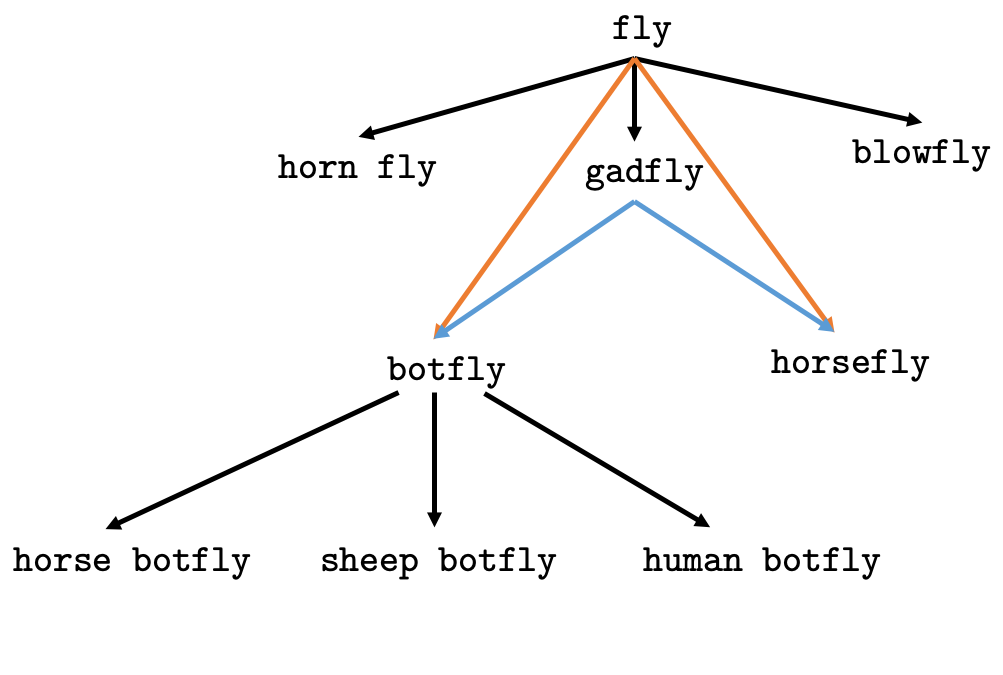}
\vspace{-0.43cm}
\caption{\textit{A fragment of a subtree from the \wordnet~ hierarchy. Orange indicates incorrectly predicted edges and blue indicates missed edges.}}
\label{fig:flat_tree}
\end{figure}

In many error cases, \model{} predicts a tree with edges that connect terms to their non-parent ancestors, skipping the direct parents. We show an example of this error in Figure \ref{fig:flat_tree}. In this fragment (taken from one of the subtrees in the dev set), the model predicts a tree in which \texttt{botfly} and \texttt{horsefly} are direct children of \texttt{fly}, bypassing the correct parent \texttt{gadfly}. On the dev set, 38.8\% of incorrect parenthood edges were cases of this type of error.

Missing edges result in predicted trees that are generally flatter than the gold tree. While all the gold trees have a height of 3 (4 nodes in the longest path from the root to any leaf), the predicted dev trees have a mean height of 2.61. Our approach scores the edges independently, without considering the structure of the tree beyond local parenthood edges. One potential way to address the bias towards flat trees is to also model the global structure of the tree (e.g., ancestor and sibling relations).

\subsection{Model Struggle Near Leaf Nodes}

\begin{table}[!th]
\centering
\begin{tabular}{|l|l|l|l|}
\hline
    & $d=1$  & $d=2$  & $d=3$    \\ \hline
$l=1$ & 81.2 & 52.3 & 39.7   \\ \hline
$l=2$ &      & 74.4 & 48.9   \\ \hline
$l=3$ &      &      & 66.0   \\ \hline
\end{tabular}
\caption{\textit{Ancestor Edge Recall, Categorized by Descendant Node Depth $d$ and Parent Edge Length $l$.} Ancestor edge prediction recall decreases with deeper descendant nodes and closer ancestor-descendant relations.}
\label{tab:depth_analysis}
\end{table}

\model{} generally makes more errors in predicting edges involving nodes that are farther from the root of each subtree. In Table \ref{tab:depth_analysis} we show the recall of ancestor edges, categorized by the number of parent edges $d$ between the subtree root and the descendant of each edge, and the number of parent edges $l$ between the ancestor and descendant of each edge. The model has lower recall for edges involving descendants that are farther from the root (higher $d$). In permutation tests of the correlation between edge recall and $d$ conditioned on $l$, 0 out of 100,000 permutations yielded a correlation at least as extreme as the observed correlation.

\subsection{Subtrees Higher Up in \wordnet~ are Harder, and Physical Entities are Easier than Abstractions}
Subtree performance also corresponds to the depth of the subtree in the entire \wordnet~ hierarchy. The $F_1$ score is positively correlated with the depth of the subtree in the full \wordnet~hierarchy, with a correlation of 0.27 (significant at p=0.004 using a permutation test with 100,000 permutations).

The subtrees included in this task span many different domains, and can be broadly categorized into subtrees representing concrete entities (such as \texttt{telephone}) and those representing abstractions (such as \texttt{sympathy}). \wordnet~ provides this categorization using the top-level synsets \texttt{physical\_entity.n.01} and \texttt{abstraction.n.06}. These categories are direct children of the root of the full \wordnet~ hierarchy (\texttt{entity.n.01}), and split almost all \wordnet~terms into two subsets. The model produces a mean $F_1$ score of 60.5 on subtrees in the \texttt{abstraction} subsection of \wordnet, and a mean $F_1$ score of 68.9 on subtrees in the \texttt{physical\_entity} subsection. A one-sided Mann-Whitney rank test shows that the model performs systematically worse on \texttt{abstraction} subtrees (compared to \texttt{physical entity} subtrees) (p=0.01).

\subsection{Pretraining Corpus Covers Most Terms}

\begin{figure}[ht]
\centering
\includegraphics[width=0.93\linewidth
]{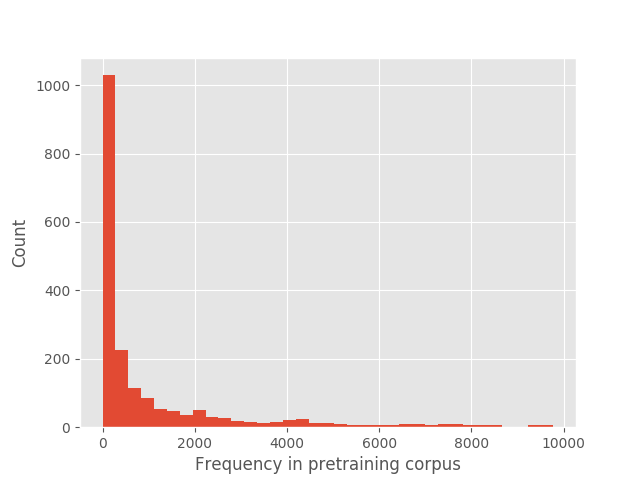}
\vspace{-0.43cm}
\caption{\textit{Frequency of terms in the \wordnet~dataset in the pretraining corpus.} Over 97\% of terms in the \citet{bansal2014structured} dataset occur at least once in the pretraining corpus. Over 80\% of terms occur less than 50k times.}
\label{fig:frequencies}
\end{figure}
With models pretrained on large web corpora, the distinction between the settings with and without access to the web at test time is less clear, since large pretrained models can be viewed as a compressed version of the web. To quantify the extent the evaluation setting measures model capability to generalize to  taxonomies consisting of unseen words, we count the number of times each term in the \wordnet~ dataset occurs in the pretraining corpus. We note that the \wordnet~glosses do not directly appear in the pretraining corpus. In Figure \ref{fig:frequencies} we show the distribution of the frequency with which the terms in the \citet{bansal2014structured} dataset occur in the \textsc{BERT} pretraining corpus.\interfootnotelinepenalty=10000 \footnote{ Since the original pretraining corpus is not available, we follow \citet{Devlin2019BERTPO} and recreate the dataset by crawling \texttt{http://smashwords.com} and Wikipedia. }  We find that  over 97\% of the terms occur at least once in the pretraining corpus. However, the majority of the terms are not very common words, with over 80\% of terms occurring less than 50k times. While this shows that the current setting does not measure model ability to generalize to completely unseen terms, we find that the model does not perform substantially worse on edges that contain terms that do not appear in the pretraining corpus. Furthermore, the model is able do well on rare terms. Future work can investigate model ability to construct taxonomies from terms that are not covered in pretraining corpora.

% \subsubsection{Term Polysemy}
% [polysemy vs performance without retrieval]
% [polysemy vs performance with retrieval]

\subsection{\wordnet{} Contains Ambiguous Subtrees}

\begin{figure}[ht]
\centering
\includegraphics[width=0.8\linewidth
]{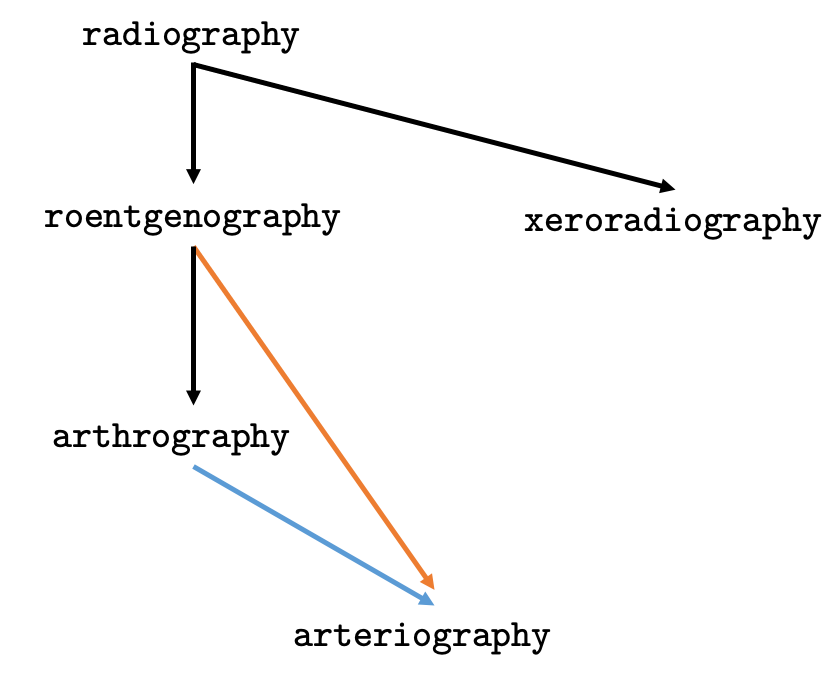}
\vspace{-0.43cm}
\caption{\textit{A fragment of a subtree from the \wordnet~ hierarchy. Orange indicates incorrectly predicted edges and blue indicates edges that were missed.}}
\label{fig:incorrect_tree}
\end{figure}

\begin{figure*}[htb]
\centering
\includegraphics[width=\textwidth
]{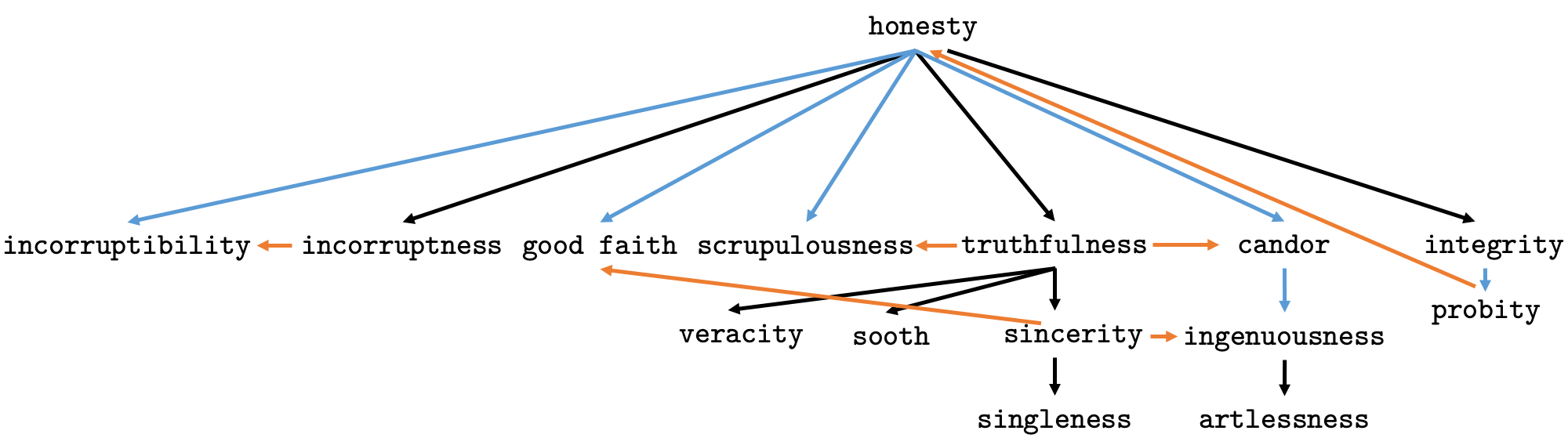}
\vspace{-0.43cm}
\caption{\textit{A fragment of a subtree from the \wordnet~ hierarchy. Orange indicates incorrectly predicted edges and blue indicates edges that were missed.}}
\label{fig:ambiguous_tree}
\end{figure*}

Some trees in the gold \wordnet~hierarchy contain ambiguous edges. Figure \ref{fig:incorrect_tree} shows one example. In this subtree, the model predicts \texttt{arteriography} as a sibling of \texttt{arthrography} rather than as its child. The definitions of these two terms suggest why the model may have considered these terms as siblings: \texttt{arteriograms} produce images of arteries while \texttt{arthrograms} produce images of the inside of joints. In Figure \ref{fig:ambiguous_tree} we show a second example of an ambiguous tree. The model predicts \texttt{good faith} as a child of \texttt{sincerity} rather than as a child of \texttt{honesty}, but the correct hypernymy relation between these terms is unclear to the authors, even after referencing multiple dictionaries.

These examples point to the potential of augmenting or improving the relations listed in \wordnet~using semi-automatic methods.

\subsection{Web-Retrieved Glosses Are Beneficial When They Contain Lexical Overlap}
We compare the predictions of \textsc{RoBERTa}-Large, with and without web glosses, to understand what kind of glosses help. We split the parenthood edges in the gold trees into two groups based on the glosses: (1) lexical overlap (the parent term appears in the child gloss and/or the child term appears in the parent gloss) and (2) no lexical overlap (neither the parent term nor the child term appears in the other term's gloss). We find that for edges in the ``lexical overlap" group, glosses increase the recall of the gold edges from 60.9 to 67.7. For edges in the ``no lexical overlap" group, retrieval decreases the recall (edge recall changes from 32.1 to 27.3).

\subsection{Pretraining and Tree Reconciliation Both Contribute to Taxonomy Construction}
We performed an ablation study in which we ablated either the pretrained language models for the parenthood prediction step or we ablated the tree reconciliation step. We ablated the pretrained language models in two ways. First, we used a one-layer LSTM on top of GloVe vectors instead of a pretrained language model as the input to the finetuning step, and then performed tree reconciliation as before. Second, we used a randomly initialized \textsc{RoBERTa}-Large model in place of a pretrained network, and then performed tree reconciliation as before. We ablated the tree reconciliation step by substituting the graph-based reconciliation step with a simpler threshold step, where we output a parenthood-relation between all pairs of words with softmax score greater than 0.5. We used the parenthood prediction scores from the fine-tuned \textsc{RoBERTa}-Large model, and substituted tree reconciliation with thresholding.

\begin{table}[!th]
\centering
%\columnwidth
\begin{tabular}{lccc}
\toprule 
& \textbf{P}  & \textbf{R}  & \textbf{F1} \\  \midrule
\textsc{RoBERTa}-Large & 71.2 & 65.9 & 67.4 \\\midrule
   \hspace{0.4cm} w/o tree reconciliation & 70.8 & 45.8 & 51.1 \\\midrule
\textsc{RoBERTa}-Random-Init & 32.6 & 28.2 & 29.3 \\\midrule
LSTM GloVe & 32.5 & 23.6 & 26.6 \\\midrule
\end{tabular}
\caption{\textit{Ablation study.} Pretraining and tree reconciliation both contribute to taxonomy construction.}
\label{tab:ablation}
\end{table}

In Table \ref{tab:ablation}, we show the results of our ablation experiments. These results show that both steps (using pretrained language models for parenthood-prediction and performing tree reconciliation) are important for taxonomy construction. Moreover, these results show that the incorporation of a new information source (knowledge learned by pretrained language models) produces the majority of the performance gains.

\subsection{Models Struggle to Generalize to Large Taxonomies}
To test generalization to large subtrees, we tested our models on the English environment and science taxonomies from SemEval-2016 Task 13 \citep{bordea2016}. Each of these taxonomies consists of a single large taxonomic tree with between 125 and 452 terms. Following \citet{mao-etal-2018-end} and \citet{shang2020taxonomy}, we used the medium-sized trees from \citet{bansal2014structured} to train our models. During training, we excluded all medium-sized trees from the \citet{bansal2014structured} dataset that overlapped with the terms in the SemEval-2016 Task 13 environment and science taxonomies.

In Table \ref{tab:semeval} we show the performance of the \textsc{RoBERTa}-Large \model~model. We show the Edge-F1 score rather than the Ancestor-F1 score in order to compare to previous work. Although the \model{} model outperforms previous work in constructing medium-sized taxonomies, this model is limited in its ability to generalize to large taxonomies. Future work can incorporate modeling of the global tree structure into \model.

\begin{table*}[h]
\centering
%\columnwidth
\begin{tabular}{llccc}
\toprule 
Dataset & Model & \textbf{P}  & \textbf{R}  & \textbf{F1} \\  \midrule
\multirow{3}{*}{Science (Averaged)} & \model & 29.4 & 28.8 & 29.1 \\
& \citet{mao-etal-2018-end} & 37.9 & 37.9 & 37.9  \\
& \citet{shang2020taxonomy} & 84.0 & 30.0 & 44.0 \\\midrule
\multirow{3}{*}{Environment (Eurovoc)} & \model & 23.1 & 23.0 & 23.0 \\
& \citet{mao-etal-2018-end} & 32.3 & 32.3 & 32.3  \\
& \citet{shang2020taxonomy} & 89.0 & 24.0 & 37.0 \\\midrule
\end{tabular}
\caption{\textit{Generalization to large taxonomic trees.} Models trained on medium-sized taxonomies generalize poorly to large taxonomies. Future work can improve the usage of global tree structure with \model.}
\label{tab:semeval}
\end{table*}

\section{Related Work}\label{sec:related_work}
%\textbf{Taxonomy Induction}
Taxonomy induction has been studied extensively, with both pattern-based and distributional approaches. Typically, taxonomy induction involves hypernym detection, the task of extracting candidate terms from corpora, and hypernym organization, the task of organizing the terms into a hierarchy. 

While we focus on hypernym organization, many systems have studied the related task of hypernym detection. Traditionally, systems have used pattern-based features such as Hearst patterns to infer hypernym relations from large corpora \citep[e.g.][]{hearst1992automatic,snow2005learning,kozareva2010semi}. For example, \citet{snow2005learning} propose a system that extracts pattern-based features from a corpus to predict hypernymy relations between terms. \citet{kozareva2010semi} propose a system that similarly uses pattern-based features to predict hypernymy relations, in addition to harvesting relevant terms and using a graph-based longest-path approach to construct a legal taxonomic tree.

Later work suggests that, for hypernymy detection tasks, pattern-based approaches outperform those based on distributional models \citep{roller-etal-2018-hearst}. Subsequent work pointed out the sparsity that exists in pattern-based features derived from corpora, and showed that combining distributional and pattern-based approaches can improve hypernymy detection by addressing this problem \citep{yu2020distributional}.

In this work we consider the task of organizing a set of terms into a medium-sized taxonomic tree. \citet{bansal2014structured} treat this as a structured learning problem and use belief propagation to incorporate siblinghood information. \citet{mao-etal-2018-end} propose a reinforcement learning based approach that combines the stages of hypernym detection and hypernym organization. In addition to the task of constructing medium-sized \wordnet~subtrees, they show that their approach can leverage global structure to construct much larger taxonomies from the SemEval-2016 Task 13 benchmark dataset, which contain hundreds of terms \citep{task13semeval2016}. \citet{shang2020taxonomy} apply graph neural networks and show that they improve performance in constructing large taxonomies in the SemEval-2016 Task 13 dataset.

%\textbf{Explicit Knowledge from Pretrained Language Models}
Another relevant line of work involves extracting structured declarative knowledge from pretrained language models. For instance, \citet{bouraoui2019inducing} showed that a wide range of relations can be extracted from pretrained language models such as \textsc{BERT}. Our work differs in that we consider tree structures and incorporate web glosses. \citet{Bosselut2019COMETCT} use pretrained models to generate explicit open-text descriptions of commonsense knowledge. Other work has focused on extracting knowledge of relations between entities \citep{petroni2019language, jiang2020can}. \citet{Blevins2020MovingDT} use a similar approach to ours for word sense disambiguation, and encode glosses with pretrained models.
\section{Discussion}\label{sec:discussion}
Our experiments show that pretrained language models can be used to construct taxonomic trees. Importantly, the knowledge encoded in these pretrained language models can be used to construct taxonomies without additional web-based information. This approach produces subtrees with higher mean $F_1$ scores than previous approaches, which used information from web queries.

When given web-retrieved glosses, pretrained language models can produce improved taxonomic trees. The gain from accessing web glosses shows that incorporating both implicit knowledge of input terms and explicit textual descriptions of knowledge is a promising way to extract relational knowledge from pretrained models. Error analyses suggest specific avenues of future work, such as improving predictions for subtrees corresponding to abstractions, or explicitly modeling the global structure of the subtrees.

Experiments on aligned multilingual \wordnet~ datasets emphasize that more work is needed in investigating the differences between taxonomic relations in different languages, and in improving pretrained language models in non-English languages. Our results provide strong baselines for future work on constructing taxonomies for different languages.
\section{Ethical Considerations}
While taxonomies (e.g., \wordnet{}) are often used as ground-truth data, they have been shown to contain offensive and discriminatory content \citep[e.g.,][]{Broughton2019TheRR}. Automatic systems created by pretrained language models can reflect and exacerbate the biases contained by their training corpora. More work is needed to detect and combat biases that arise when constructing and evaluating taxonomies.

Furthermore, we used previously constructed alignments to extend our results to wordnets in multiple languages. While considering English \wordnet~as the basis for the alignments allows for convenient comparisons between languages and is the standard method for aligning wordnets across languages, continued use of these alignments to evaluate taxonomy construction imparts undue bias towards conceptual relations found in English.
\section{Acknowledgements}
We thank the members of the Berkeley NLP group and the anonymous reviewers for their insightful feedback. CC and KL are supported by National Science Foundation Graduate Research Fellowships. This research has been supported by DARPA under agreement HR00112020054. The content does not necessarily reflect the position or the policy of the government, and no official endorsement should be inferred.

\bibliographystyle{acl_natbib}
\bibliography{naacl2021}
\clearpage\newpage
\appendix
\section*{Appendix}
\subsection*{Language-Specific Pretrained Models}\label{sec:mling_models_list}
We used pretrained models from the following sources: \newline \url{https://github.com/codegram/calbert},\newline \url{https://github.com/google-research/bert/blob/master/multilingual.md} \citep{Devlin2019BERTPO},\newline \url{http://turkunlp.org/FinBERT/} \citep{finnishbert},\newline \url{https://github.com/dbmdz/berts},\newline \url{https://github.com/wietsedv/bertje} \citep{dutchbert},\newline \url{https://huggingface.co/dkleczek/bert-base-polish-uncased-v1},\newline \url{https://github.com/neuralmind-ai/portuguese-bert},\newline \url{https://github.com/dccuchile/beto/blob/master/README.md} \citep{spanishbert}

\subsection*{Multilingual \wordnet~Dataset Statistics}
Table \ref{tab:multilingual_dataset} details the datasets we created by using synset alignments to the English dataset proposed in \citet{bansal2014structured}. The data construction method is described in Section \ref{sec:datasets}.
\begin{table}[!h]
\centering
%\columnwidth
\begin{tabular}{lrrrrrr}
\toprule
~& \multicolumn{3}{c}{\textbf{Num}} & \multicolumn{3}{c}{\textbf{Average}} \\
~& \multicolumn{3}{c}{\textbf{Trees}} & \multicolumn{3}{c}{\textbf{Nodes per Tree}}\\
~& Train & Dev & Test & Train & Dev & Test \\  \midrule
 ca & 391 & 94 & 90 & 9.2 & 9.3 & 8.7\\  \midrule
 zh & 216 & 48 & 64 & 10.0 & 12.4 & 9.2\\  \midrule
 en & 533 & 114 & 114 & 19.7 & 20.3 & 19.8\\  \midrule
 fi & 532 & 114 & 114 & 17.8 & 18.8 & 18.1\\  \midrule
 fr & 387 & 82 & 76 & 8.7 & 9.1 & 8.3\\  \midrule
 it & 340 & 85 & 75 & 6.3 & 7.2 & 6.2\\  \midrule
 nl & 308 & 58 & 64 & 6.6 & 6.7 & 6.3\\  \midrule
 pl & 283 & 73 & 72 & 7.7 & 8.0 & 7.4\\  \midrule
 pt & 347 & 68 & 77 & 7.1 & 8.2 & 7.2\\  \midrule
 es & 280 & 60 & 60 & 6.5 & 6.1 & 5.8\\ \bottomrule
\end{tabular}
\caption{\textit{Dataset Statisics.} 
For each language, we show the number of train, dev, and test subtrees that remain after the subsetting procedure described in Section \ref{sec:multilingual_dataset}. In addition, we show the mean number of nodes per tree in each language. We use ISO 639-1 language acronyms.}\label{tab:multilingual_dataset}
\end{table}

\subsection*{Ablation Results}
Table \ref{tab:ablation_learning_rates} shows the results for the learning rate trials for the ablation experiment.
\begin{table}[!h]
\centering
%\columnwidth
\begin{tabular}{lccc}
\toprule 
& \textbf{1e-5}  & \textbf{1e-6}  & \textbf{1e-7} \\  \midrule
\textsc{RoBERTa}-Large & 59.5 & \textbf{67.3} & 60.7 \\\midrule
\hspace{0.10cm}w/o tree reconciliation & 38.6 & \textbf{51.2} & 18.2 \\\midrule
\textsc{RoBERTa}-Random-Init & 17.4 & 26.4 & \textbf{27.0}  \\\midrule
\end{tabular}
\caption{\textit{Dev F1 Scores for Different Learning Rates, Ablation Experiments}
.}
\label{tab:ablation_learning_rates}
\end{table}

Table \ref{tab:ablation_trials} shows the results for the test trials for the ablation experiment.
\begin{table}[!h]
\centering
%\columnwidth
\begin{tabular}{lccc}
\toprule 
& \textbf{Run 0}  & \textbf{Run 1}  & \textbf{Run 2} \\  \midrule
\textsc{RoBERTa}-Large & 67.1 & 67.3 & 67.7 \\\midrule
\hspace{0.10cm}w/o tree reconciliation & 51.2 & 51.4 & 50.6 \\\midrule
\textsc{RoBERTa}-Random-Init & 27.0 & 29.9 & 31.1 \\\midrule
LSTM GloVe & 24.6 & 27.7 & 27.6 \\\midrule
\end{tabular}
\caption{\textit{Dev F1 Scores for Three Trials, Ablation Experiments}
.}
\label{tab:ablation_trials}
\end{table}

\subsection*{SemEval Results}
\begin{table}[!h]
\centering
%\columnwidth
\begin{tabular}{llccc}
\toprule 
Dataset & \textbf{Run 0}  & \textbf{Run 1}  & \textbf{Run 2} \\  \midrule
Science (Combined) & 28.6 & 31.7 & 25.1 \\
Science (Eurovoc) & 26.6 & 37.1 & 31.5 \\
Science (WordNet) & 26.5 & 28.8 & 25.8 \\
Environment (Eurovoc) & 23.4 & 21.5 & 24.2 \\\midrule
\end{tabular}
\caption{\textit{Test F1 Scores for Three Trials, Semeval.} We show the Edge-F1 score rather than the Ancestor-F1 score in order to compare to previous work.}
\label{tab:semeval_trials}
\end{table}

Table \ref{tab:semeval_trials} shows the results for the test trials for the SemEval experiment. These results all use the \textsc{RoBERTa}-Large model in the parenthood prediction step.

\subsection*{Random Baseline for Multilingual \wordnet~Datasets}\label{sec:random_baseline}
To compute the random baseline in each language, we randomly construct a tree containing the nodes in each test tree and compute the ancestor precision, recall and $F_1$ score on the randomly constructed trees. We include the $F_1$ scores for three trials in Table \ref{tab:baseline_trials}.
\begin{table}[!ht]
\centering
\begin{tabular}{lccc}
\toprule
    Model & \textbf{Run 0} & \textbf{Run 1} & \textbf{Run 2} \\\midrule
    Catalan & 19.7 & 19.1 & 21.2 \\\midrule
    Chinese & 23.5 & 26.8 & 27.0 \\\midrule
    English & 8.1 & 8.9 & 9.7 \\\midrule
    Finnish & 10.6 & 10.0 & 9.8 \\\midrule
    French & 22.1 & 24.7 & 19.4 \\\midrule
    Italian & 28.0 & 27.1 & 31.6 \\\midrule
    Dutch & 29.7 & 27.9 & 22.8 \\\midrule
    Polish & 20.5 & 22.1 & 27.5 \\\midrule
    Portuguese & 27.9 & 28.1 & 22.2 \\\midrule
    Spanish & 32.6 & 24.1 & 24.3 \\\bottomrule
\end{tabular}
\caption{\textit{Test F1 Scores for Three Trials Using a Random Baseline.}
}\label{tab:baseline_trials}
\end{table}

\subsection*{Subtree Construction Results, English WordNet}
Table \ref{tab:en_lr} shows the results for the learning rate trials for the English \wordnet~experiment.
\begin{table}[!h]
\centering
\begin{tabular}{lccc}
\toprule
     Model & 1e-5 & 1e-6 & 1e-7 \\\midrule
    \textsc{BERT} & 60.0 & \textbf{63.3} & 60.7 \\\midrule
    \textsc{BERT}-Large & 59.5 & \textbf{67.3} & 65.8 \\\midrule
    \textsc{RoBERTa}-Large & 56.3 & \textbf{67.1} & 65.5 \\\midrule
    \textsc{RoBERTa}-Large\\(Web-retrieved Glosses) & 58.6 & \textbf{71.5} & 64.7 \\\midrule
    \textsc{RoBERTa} Large\\(WordNet Glosses) & 63.0 & \textbf{83.7} & 82.9 \\\bottomrule
\end{tabular}
\caption{\textit{Dev Results for Different Learning Rates, English Models. We highlight in bold the best learning rate for each model.} }\label{tab:en_lr}
\end{table}

Table \ref{tab:en_test_trials} shows the results for the test trials for the English \wordnet~experiment.
\begin{table}[!h]
\centering
\begin{tabular}{lccc}
\toprule
    Model & \textbf{Run 0} & \textbf{Run 1} & \textbf{Run 2} \\\midrule
    \textsc{BERT} & 53.6 & 54.0 & 52.5 \\\midrule
    \textsc{BERT}-Large & 58.9 & 61.5 & 63.8 \\\midrule
    \textsc{RoBERTa}-Large & 62.9 & 64.2 & 63.3 \\\midrule
    \textsc{RoBERTa}-Large\\(Web-retrieved\\glosses) & 66.6 & 66.3 & 67.1 \\\midrule
    \textsc{RoBERTa}-Large\\(WordNet glosses) & 82.4 & 84.0 & 83.2 \\\bottomrule
\end{tabular}
\caption{\textit{Test F1 Scores for Three Trials, English.} }\label{tab:en_test_trials}
\end{table}

\subsection*{Subtree Construction Results, Multilingual WordNet}
Table \ref{tab:multilingual_lr} shows the results for the learning rate trials for the non-English \wordnet~experiments.
\begin{table}[!h]
\centering
\begin{tabular}{llccc}
\toprule
    Language & Model & 1e-5 & 1e-6 & 1e-7 \\\midrule
    \multirow{2}{*}{Catalan} & Calbert & \textbf{39.9} & 37.9 & 24.5 \\
     & mBERT & 39.7 & \textbf{43.5} & 32.6 \\\midrule
    \multirow{2}{*}{Chinese} & Chinese BERT & 56.9 & \textbf{59.0} & 54.3 \\
     & mBERT & 57.4 & \textbf{60.6} & 44.7 \\\midrule
    \multirow{2}{*}{Finnish} & FinBERT & 45.6 & \textbf{50.1} & 47.0 \\
     & mBERT & 24.6 & \textbf{30.2} & 28.9 \\\midrule
    \multirow{2}{*}{French} & French BERT & 48.9 & \textbf{50.6} & 46.9 \\
     & mBERT & 40.3 & \textbf{41.1} & 32.5 \\\midrule
    \multirow{2}{*}{Italian} & Italian BERT & \textbf{52.6} & 52.2 & 46.9 \\
     & mBERT & 50.7 & \textbf{51.8} & 41.3 \\\midrule
    \multirow{2}{*}{Dutch} & BERTje & \textbf{49.0} & 48.8 & 38.1 \\
     & mBERT & \textbf{44.9} & 44.5 & 32.9 \\\midrule
    \multirow{2}{*}{Polish} & Polbert & \textbf{54.2} & 52.9 & 48.2 \\
     & mBERT & \textbf{53.0} & 50.7 & 36.4 \\\midrule
    \multirow{2}{*}{Portuguese} & BERTimbau & 51.2 & \textbf{52.0} & 42.1 \\
     & mBERT & \textbf{38.5} & 37.8 & 28.0 \\\midrule
    \multirow{2}{*}{Spanish} & BETO & 56.7 & \textbf{57.4} & 52.8 \\
     & mBERT & \textbf{49.5} & 41.5 & 40.4 \\\bottomrule
\end{tabular}
\caption{\textit{Dev Results for Different Learning Rates, Multilingual. We highlight in bold the best learning rate for each model.} }\label{tab:multilingual_lr}
\end{table}
% Best epoch after training for 15 epochs each.

Table \ref{tab:multilingual_test_trials} shows the results for the test trials for the non-English \wordnet~experiments.
\begin{table*}[!h]
\centering
\begin{tabular}{llccc}
\toprule
    Language & Model & \textbf{Run 0} & \textbf{Run 1} & \textbf{Run 2} \\\midrule
    \multirow{2}{*}{Catalan} & Calbert & 36.5 & 34.1 & 33.6 \\
     & mBERT & 39.4 & 41.8 & 32.7 \\\midrule
    \multirow{2}{*}{Chinese} & Chinese BERT & 57.1 & 62.3 & 56.8 \\
     & mBERT & 55.2 & 59.4 & 58.0 \\\midrule
    \multirow{2}{*}{Finnish} & FinBERT & 43.6 & 44.6 & 43.2 \\
     & mBERT & 25.5 & 26.3 & 26.7 \\\midrule
    \multirow{2}{*}{French} & French BERT & 47.5 & 49.5 & 50.4 \\
     & mBERT & 41.0 & 40.9 & 38.9 \\\midrule
    \multirow{2}{*}{Italian} & Italian BERT & 43.2 & 47.2 & 47.8 \\
     & mBERT & 42.9 & 43.6 & 49.3 \\\midrule
    \multirow{2}{*}{Dutch} & BERTje & 43.8 & 44.9 & 42.4 \\
     & mBERT & 35.9 & 33.0 & 27.1 \\\midrule
    \multirow{2}{*}{Polish} & Polbert & 51.2 & 49.9 & 47.3 \\
     & mBERT & 40.1 & 42.0 & 41.5 \\\midrule
    \multirow{2}{*}{Portuguese} & BERTimbau & 57.6 & 57.4 & 55.8 \\
     & mBERT & 38.4 & 38.2 & 34.3 \\\midrule
    \multirow{2}{*}{Spanish} & BETO & 50.8 & 53.4 & 50.9 \\
     & mBERT & 48.7 & 49.3 & 44.0 \\\bottomrule
\end{tabular}
\caption{\textit{Test F1 Scores for Three Trials, Multilingual.} }\label{tab:multilingual_test_trials}
\end{table*}

Table \ref{tab:multilingual_full} shows the results for all tested models for the non-English \wordnet~experiments.
\begin{table*}[!h]
\centering
\begin{tabular}{llccc}
\toprule
    Language & Model & \textbf{P} & \textbf{R} & \textbf{F1} \\\midrule
    \multirow{2}{*}{Catalan} & Calbert & 39.3 & 32.4 & 34.7 \\
     & mBERT & 38.7 & 39.7 & 38.0 \\\midrule
    \multirow{2}{*}{Chinese} & Chinese BERT & 62.2 & 57.3 & 58.7 \\
     & mBERT & 61.9 & 56.0 & 57.5 \\\midrule
    \multirow{2}{*}{Finnish} & FinBERT & 47.9 & 42.6 & 43.8 \\
     & mBERT & 29.6 & 25.4 & 26.2 \\\midrule
    \multirow{2}{*}{French} & French BERT & 51.3 & 49.1 & 49.1 \\
     & mBERT & 43.3 & 40.0 & 40.3 \\\midrule
    \multirow{2}{*}{Italian} & Italian BERT & 48.3 & 45.5 & 46.1 \\
     & mBERT & 47.6 & 44.6 & 45.3 \\\midrule
    \multirow{2}{*}{Dutch} & BERTje & 44.6 & 44.8 & 43.7 \\
     & mBERT & 34.3 & 31.6 & 32.0 \\\midrule
    \multirow{2}{*}{Polish} & Polbert & 51.9 & 49.7 & 49.5 \\
     & mBERT & 43.7 & 41.4 & 41.2 \\\midrule
    \multirow{2}{*}{Portuguese} & BERTimbau & 59.3 & 57.1 & 56.9 \\
     & mBERT & 38.7 & 38.2 & 37.0 \\\midrule
    \multirow{2}{*}{Spanish} & BETO & 53.1 & 51.7 & 51.7 \\
     & mBERT & 47.3 & 49.4 & 47.3 \\\bottomrule
\end{tabular}
\caption{\textit{Multilingual \wordnet~Test Results.} We use ISO 639-1 acronyms to indicate languages.}\label{tab:multilingual_full}
\end{table*}

\end{document}